\documentclass[mlabstract,onecolumn]{jmlr}

\jmlrproceedings{PMLR}{Extended Abstract Track}

\usepackage{booktabs}

\jmlrvolume{}
\firstpageno{1}
\editors{}
\jmlryear{2026}
\jmlrworkshop{Symmetry and Geometry in Neural Representations}
\newcommand{\rel}{\mathrm{rel}\,\ell_2}

\title[Predicting Cable Dynamics with Physical Attention Bias]{Predicting Cable Dynamics\titlebreak with Physical Attention Bias}

\author{\Name{Avihai Giuili\nametag{\thanks{Equal contribution.}}} \Email{avigiuili@mail.tau.ac.il}\\
  \addr Tel Aviv University
  \AND
  \Name{Rotem Atari\nametag{\footnotemark[1]}} \Email{rotematari@mail.tau.ac.il}\\
  \addr Tel Aviv University
  \AND
  \Name{Avishai Sintov} \Email{sintov1@tauex.tau.ac.il}\\
  \addr Tel Aviv University
  \AND
  \Name{Maya Bechler-Speicher} \Email{mayabs@meta.com}\\
  \addr Meta AI}

\begin{document}

\maketitle

\begin{abstract}
Learned simulators for deformable linear objects (DLOs) such as cables have to
predict the motion of cables they were not trained on and stay stable over long
rollouts. Most of their error occurs where the cable touches itself or the
floor. Attention over all pairs of cable segments can represent contact
between parts of the cable that are far apart along its length, but attention
has no notion of geometry. A cable has two pairwise distances, which agree
only while it is straight: the arc-length distance along the cable, which
governs elastic forces, and the Euclidean distance in space, which governs
contact. We add a \emph{physical attention bias}, an additive term on the
attention logits with a learned rate, and ask which distance it should use.
We compare no bias, each distance alone, and both distances on disjoint sets
of heads, keeping the rest of the model and the training protocol fixed. A
physical bias improves prediction on unseen cables. The gain is largest when
attention is the only mechanism that connects distant segments: there, the
arc-length bias reduces prediction error by $15\%$ and more than halves the
drift in segment length. The Euclidean bias alone stays close to unbiased
attention, while assigning both distances across heads is best or near-best on
every metric we report. Code and per-run records: \url{https://github.com/avihaig/dlogps}.
\end{abstract}


\section{Introduction}
\label{sec:intro}

A robot that manipulates a deformable linear object (DLO) such as a cable,
rope, or suture needs a model of how it will move
\citep{caporali2026dlo_survey,yu2025whole_body}. Rod mechanics gives one
\citep{bergou2008der}, but it is expensive to solve and needs material
parameters that are hard to identify for a new cable. Learned dynamics models instead fit trajectories and are cheap to run online
\citep{yang2021inbilstm,gu2025ganet}. They have to generalize in two ways. The cable met at deployment is not a training cable, and the model is
trained on single steps but run in closed loop for hundreds of them, so its
own errors become its next inputs.

Graph-based learned simulators update a graph by message passing, and the
edge set decides which nodes interact. GNS rebuilds edges from a spatial
radius at every step \citep{sanchezgonzalez2020gns}, and MeshGraphNets adds
world edges between nodes that are close in space but not connected in the
mesh \citep{pfaff2021meshgraphnets}. For a DLO this choice matters. A cable is
a one-dimensional chain in $\mathbb{R}^3$ and so has two distances: the
\emph{intrinsic} arc-length distance along the chain, which governs elastic
forces, and the \emph{extrinsic} Euclidean distance, which governs
self-contact between segments that are close in space but far apart along the
chain. The two agree while the cable is straight and diverge once it folds,
and folding is where rollouts accumulate most of their error.

Hybrid graph transformers move this choice from the edge set into attention.
A GraphGPS layer runs local message passing in parallel with dense attention
over all node pairs \citep{rampasek2022graphgps}, so either distance can enter
the attention branch as a pairwise term without changing the graph. Graph
transformers already add structure this way, through additive logit biases or
attention masks, but mostly on static benchmarks and usually combining several
signals \citep{ying2021graphormer,luo2023transformerm,%
liu2024gradformer,yun2026hopformer}. The closest hybrid model for DLOs
aggregates local interactions before a Transformer encoder, for quasi-static
2D shape control \citep{yu2025eapegat}. Concurrent work on a cable suspended
from a drone scales all attention logits by one state-dependent scalar
computed from learned Lagrangian matrices \citep{pierallini2026pitr}; this
changes how sharp attention is but adds no pairwise structure. So it is still
open which distance should bias global attention during a rollout, and
whether the answer changes when the local stream is removed.

This extended abstract is a first study of that question. We change only the
pairwise distance and keep the backbone, optimizer, seeds, and evaluation
windows fixed. We compare five variants with the same parameter budget, two
of which are \emph{dual-metric}: they put the two distances on disjoint
subsets of heads. We evaluate each variant with the local stream on and off,
and compare against chain-only, per-node MLP, and global LSTM baselines.

\section{Method}
\label{sec:method}

\paragraph{Representation and target.}
The cable is $N{=}33$ vertices joined by $N{-}1$ bidirectional chain edges.
Node features are a $C{=}70$-frame velocity history, four material parameters,
and the clipped floor distance; edge features are relative displacement,
distance, and rest length. Absolute coordinates are never a feature, so the
representation is invariant to translations parallel to the floor. The network
predicts a per-node displacement; training is one-step with random-walk input
noise \citep{sanchezgonzalez2020gns}, evaluation autoregressive.

\paragraph{Hybrid block.}
Linear encoders lift node and edge features to width $d$, $L$ identical layers
follow, and a linear readout maps node embeddings to displacements. Each layer
runs two streams on the \emph{same} input, in parallel
\citep{rampasek2022graphgps}: a GatedGCN over the chain edges, which alone
reads and updates edge embeddings, and dense multi-head attention over all $N$
vertices, which sees node embeddings only. Each stream output is added to the
layer input and normalized per node; the sum passes through a two-layer MLP
with a second residual. Switching the GatedGCN off leaves global attention
alone (\emph{local off}); switching attention off leaves chain-only message
passing (\emph{chain only}).

\paragraph{Physical attention bias.}
We add one term to the attention logits, with a learned decay rate per head;
the variants differ only in \emph{which distance each head receives}.
For head $h$ the score and the attention weight are
$s^h_{ij}=q_i^h\!\cdot k_j^h/\sqrt{d_k}-b_h(i,j)$, so after the softmax the
bias acts multiplicatively,
$\alpha^h_{ij}\propto\exp(q_i^h\!\cdot k_j^h/\sqrt{d_k})\,e^{-b_h(i,j)}$:
$b_h{=}0$ recovers standard attention and a positive bias attenuates the
contribution of vertex $j$ to vertex $i$. Both metrics enter normalized to
$[0,1]$, $\widetilde C_{ij}=|i-j|/(N-1)$ and
$\widetilde D_{ij}=\|p_i-p_j\|/L$ with $L$ the rest length, so their rates are
comparable at initialization. The five variants set $b_h(i,j)$ to:
\emph{Unbiased}, $0$ on every head; \emph{Euclidean},
$\gamma_h\widetilde D_{ij}$ on every head; \emph{Chain},
$\beta_h\widetilde C_{ij}$ on every head; \emph{Mixed}, $\lfloor H/3\rfloor$
heads Euclidean, $\lfloor H/3\rfloor$ chain, the rest unbiased; and
\emph{Euclidean+Chain only}, half the heads Euclidean and half chain, none
unbiased. The rates $\gamma_h,\beta_h=\operatorname{softplus}(\theta_h)>0$ are
one learned scalar per biased head, initialized to one and shared across
layers, so a prior can be softened or sharpened but not reversed.
During rollout $\widetilde C$ is fixed by the chain while $\widetilde D$ is
recomputed from the model's \emph{own} predicted positions, so the extrinsic
bias sees the model's errors while the intrinsic bias cannot drift. Mixed and
Euclidean+Chain only are the two \emph{dual-metric} allocations, identical
except in whether the remaining heads stay unbiased. No variant adds a third distance or mixes both in one head; all differ by at most $H$ scalars.

\section{Experimental protocol}
\label{sec:protocol}

\paragraph{Data.}
Cables are simulated in MuJoCo 3.9.0 \citep{todorov2012mujoco}: two grippers
drive the cable to a randomized pose and release it, and the fall onto the
floor is recorded at $500$\,Hz until it settles. Training
uses $46$ cables spanning rest length $0.80$--$1.60$\,m, diameter
$1.5$--$10$\,mm, and five Young's moduli from $10^{6}$ to $10^{9}$\,Pa, $100$
episodes each; the $40$ test cables are interpolated inside the same hull and
never appear in training, so the model must generalize to unseen cable
instances and to long rollouts (Appendix~\ref{apd:data}).

\paragraph{Training, rollouts, metrics.}
Every model trains for $100$k steps at batch $512$ with Adam at $d{=}128$,
$H{=}8$, $L{=}4$ ($891$k parameters); the last checkpoint is reported and
every hyperparameter was chosen on validation windows, never on test cables
(Appendix~\ref{apd:training}). Each model is rolled out for
$400$ predicted frames from the same $400$ test windows. With
$\hat p_i(t)$ and $p_i(t)$ the predicted and recorded positions at predicted
frame $t$, the primary error is
$\rel(t)=\tfrac{1}{NL}\sum_i\|\hat p_i(t)-p_i(t)\|_2$, averaged over the
rollout. Two physical violations use the prediction alone: the link-length
drift $\delta$ from the rest length $\ell_0=L/(N-1)$, and the
self-intersection rate $v$, the fraction of frames in which two non-adjacent
segments pass closer than the cable diameter. A predicted frame is \emph{floor
contact} when the recorded cable has a vertex at or below one diameter
(Appendix~\ref{apd:metrics}).

\paragraph{Comparison design.}
The matrix is the five variants at three seeds with the local stream on and
off, plus three baselines at three seeds: $39$ runs. The per-node MLP never
mixes vertices; the global LSTM reads the flattened state as a $C$-step
sequence and writes all displacements. A cell reports the mean of the three
seed means $\pm$ the mean of the three window sds $\pm$ the sd of the seed
means; we do not run significance tests.

\section{Results}
\label{sec:results}

\begin{table}[t]
\centering\footnotesize
\setlength{\tabcolsep}{4pt}
\renewcommand{\arraystretch}{0.93}
\begin{tabular}{@{}lccc@{}}
\toprule
& $\rel$ (frac.\ of $L$) & $\delta$ (frac.\ of $\ell_0$) & $v$ (frac.\ of frames) \\
\midrule
\multicolumn{4}{@{}l}{\emph{local stream on}} \\
Unbiased & $0.0325\pm0.0281\pm0.0058$ & $0.0300\pm0.0159\pm0.0032$ & $0.171\pm0.231\pm0.009$ \\
Euclidean & $0.0320\pm0.0267\pm0.0042$ & $0.0303\pm0.0135\pm0.0021$ & $0.174\pm0.229\pm0.032$ \\
Chain & $0.0316\pm0.0341\pm0.0057$ & $0.0271\pm0.0124\pm0.0023$ & $0.167\pm0.227\pm0.030$ \\
Mixed & $0.0302\pm0.0251\pm0.0067$ & $\mathbf{0.0267\pm0.0109\pm0.0030}$ & $0.168\pm0.229\pm0.012$ \\
Euclidean+Chain only & $\mathbf{0.0297\pm0.0288\pm0.0035}$ & $0.0275\pm0.0129\pm0.0007$ & $\mathbf{0.165\pm0.229\pm0.012}$ \\
\midrule
\multicolumn{4}{@{}l}{\emph{local stream off}} \\
Unbiased & $0.0399\pm0.0246\pm0.0002$ & $0.1693\pm0.1891\pm0.0030$ & $0.216\pm0.258\pm0.006$ \\
Euclidean & $0.0382\pm0.0279\pm0.0007$ & $0.1470\pm0.2232\pm0.0178$ & $0.208\pm0.262\pm0.020$ \\
Chain & $\mathbf{0.0339\pm0.0195\pm0.0008}$ & $0.0783\pm0.0382\pm0.0040$ & $0.174\pm0.241\pm0.004$ \\
Mixed & $0.0352\pm0.0211\pm0.0016$ & $\mathbf{0.0779\pm0.0455\pm0.0036}$ & $0.172\pm0.242\pm0.016$ \\
Euclidean+Chain only & $0.0343\pm0.0202\pm0.0002$ & $0.0784\pm0.0409\pm0.0013$ & $\mathbf{0.161\pm0.234\pm0.006}$ \\
\midrule
\multicolumn{4}{@{}l}{\emph{baselines}} \\
chain only & $0.0925\pm0.0838\pm0.0720$ & $0.0812\pm0.0589\pm0.0322$ & $0.115\pm0.191\pm0.024$ \\
MLP & $0.0868\pm0.0785\pm0.0163$ & $1.35\pm2.32\pm0.73$ & $0.435\pm0.256\pm0.010$ \\
LSTM & $0.1070\pm0.0911\pm0.0397$ & $0.2819\pm0.3124\pm0.0509$ & $0.082\pm0.150\pm0.031$ \\
\bottomrule
\end{tabular}
\caption{Unseen-cable test, $400$ windows of $400$-frame rollouts; mean of
seed means $\pm$ mean of window sds $\pm$ sd of seed means; bold: lowest mean
per block.}
\label{tab:aggregate}
\end{table}

The largest effect in \tableref{tab:aggregate} comes from the architecture,
not the bias. Every GPS row lowers mean $\rel$ relative to chain-only message
passing, by $65$--$68\%$ with the local stream on and $57$--$63\%$ with it off,
and the MLP and LSTM baselines are far behind as well. The two streams also
fail in different ways. Removing the local stream raises $\rel$ by $7$--$23\%$
but multiplies drift by up to $5.6\times$, which suggests that dense attention
provides the long-range coupling and local message passing holds the rest
length.

Every physical bias beats unbiased attention on $\rel$ in both blocks, by
more when attention is the only path between distant segments. With the local stream off, Chain reaches
$0.0339\pm0.0008$ against $0.0399\pm0.0002$, a $15\%$ reduction with seed
means that do not overlap, and it more than halves drift. With the local
stream on, the largest gap shrinks to $9\%$ and lies inside the window spread,
so with three seeds it is only an ordering. The two
distances are not interchangeable. With the local stream off, the extrinsic
bias alone lowers $\rel$ by $4\%$ and the intrinsic bias by $15\%$, and the
intrinsic bias alone closes about two thirds of the drift gap that the whole
GatedGCN stream closes.

Splitting the local-on rollouts by phase (\tableref{tab:regimes}) shows where
the error is: $\rel$ during floor contact is about eight times that in free
flight. No single distance is best everywhere. Chain is only third on
local-on $\rel$ and worst in both phase columns, and Euclidean stays close to
unbiased whenever the local stream is off. Euclidean+Chain only, by contrast,
is lowest or within one seed standard deviation of the lowest in all six
cells of \tableref{tab:aggregate}, with Mixed close behind. We therefore suggest the dual-metric bias as a default.

\section{Conclusion}
\label{sec:conclusion}

A physical attention bias costs at most $H$ learned scalars and closes part
of the gap between the two streams: the intrinsic distance alone recovers
about two thirds of the drift benefit of the whole local stream. The effect
clearly exceeds seed noise only when attention is the sole path between
distant segments. Our results tentatively favor the dual-metric bias, with
both distances spread across heads and no head left unbiased. Caveats: the
local-on ranking lies inside the window spread, test cables lie inside the
training range, and a lowest mean does not by itself show that heads
specialize.

\bibliography{references}

\appendix

\section{Data generation}
\label{apd:data}

Each cable is a 32-segment chain ($N{=}33$ vertices) under MuJoCo's cable
elasticity plugin with twist-to-bend ratio $G/E{=}0.38$, integrated with
\texttt{implicitfast} at $\Delta t{=}4.2\times10^{-4}$\,s; kinematics are
recorded at $500$\,Hz (the actual rate is rounded to a whole number of
simulator steps, so the time step is read from the recorded timestamps). The scene is
a floor and two grippers welded to the cable ends. Each episode settles the
cable, drives the grippers quasi-statically to a randomized pose (end
separation between $0.10$\,m and a quarter of the rest length, a $0.5$\,m
sampling box, a $\pi/4$ end-facing cone, a $25$\,s reach and a $5$\,s dwell),
and releases both welds on the same tick; $t{=}0$ is that instant. Recording
ends when the 95th percentile of vertex speed stays below $1$\,cm/s for
$0.5$\,s, or after $5$\,s. Self-collision is permitted and common.

The training population is $46$ cables in five material classes (Young's
modulus $10^{6}$, $10^{7}$, $10^{8}$, $5\times10^{8}$, $10^{9}$\,Pa, each with
its own density and joint damping), four rest lengths ($0.80$, $1.10$, $1.40$,
$1.60$\,m), and diameters from $1.5$ to $10$\,mm, each released from $100$
randomized poses (two sets of $50$, drawn with different seeds). The test population is $40$
cables with interpolated lengths and diameters inside the same hull, $30$
episodes each. Test windows start at a stride of the rollout length inside
every episode; a stratified cap of $400$ windows is taken round-robin across
cables so that all models are scored on the same windows.

\section{Architecture and training}
\label{apd:training}

Node features are $F=3C+5=215$ wide at $C{=}70$: the velocity history newest
first, the four material parameters, and the floor distance clipped at $1$\,m.
The material channels are encoded as $(\log_{10}x-c)/h$ with $(c,h)$ equal to
$(7,3)$, $(-3,2)$, $(-2.301,1)$, and $(-1.3495,2.6505)$ for Young's modulus,
joint damping, diameter, and linear density; velocity history, floor distance,
edge features, and the displacement targets are standardized with statistics
fitted on the training split before noise is added and frozen thereafter;
positions are not scaled. The block uses $d{=}128$, $H{=}8$, $L{=}4$, a
two-layer ReLU MLP of hidden width $2d$, and post-residual LayerNorm over the
feature axis of each node. The GatedGCN stream normalizes its aggregation by
the gate sum. The loss is the mean squared error on the standardized one-step
displacement with random-walk velocity noise of scale $\sigma{=}0.1$\,m/s per
axis at the newest frame. Adam runs $100$k steps at batch $512$ with an
exponential decay from $10^{-3}$ to $10^{-4}$, gradient clipping at $1.0$,
weight decay $0$, and dropout $0$; the reported checkpoint is the last one for every model and seed. The per-node MLP has width $2048$ and three layers; the
global LSTM has width $256$ and three layers and receives centroid-relative
positions divided by the rest length in addition to the node features. One
run trains and is evaluated in under an hour on a single consumer GPU; the $39$ runs
total roughly $30$ GPU-hours.

\section{Metric definitions and the phase split}
\label{apd:metrics}

\begin{table}[ht]
\centering\footnotesize
\setlength{\tabcolsep}{6pt}
\begin{tabular}{@{}lcc@{}}
\toprule
& floor contact $\rel$ & free flight $\rel$ \\
\midrule
Unbiased & $0.0511\pm0.0536\pm0.0105$ & $0.0062\pm0.0095\pm0.0001$ \\
Euclidean & $0.0510\pm0.0615\pm0.0080$ & $0.0062\pm0.0099\pm0.0005$ \\
Chain & $0.0529\pm0.0871\pm0.0122$ & $0.0064\pm0.0131\pm0.0014$ \\
Mixed & $0.0484\pm0.0551\pm0.0135$ & $\mathbf{0.0058\pm0.0097\pm0.0011}$ \\
Euclidean+Chain only & $\mathbf{0.0479\pm0.0643\pm0.0068}$ & $0.0061\pm0.0127\pm0.0010$ \\
\bottomrule
\end{tabular}
\caption{Local-on rows of \tableref{tab:aggregate} split by phase of the
recorded trajectory ($400$ and $345$ eligible windows); bold: lowest mean per
column. Every column ordering here sits inside the seed spread.}
\label{tab:regimes}
\end{table}

Let $\hat p_i(t)$ and $p_i(t)$ be the predicted and recorded positions of
vertex $i$ at predicted frame $t\in\{1,\dots,400\}$ (frame $0$ is the observed
input state and is never scored) and $L$ the sample's rest length. With
$e_i(t)=\|\hat p_i(t)-p_i(t)\|_2$ and $\ell_2(t)=\frac1N\sum_i e_i(t)$,
$\rel(t)=\ell_2(t)/L$, and a rollout reports the mean of $\rel(t)$ over its
predicted frames. Link-length drift and self-intersection use the prediction
and the cable geometry only; recorded rollouts give the simulator's own
baseline on both. With $\ell_0=L/(N-1)$ and
$\hat\ell_k(t)=\|\hat p_{k+1}(t)-\hat p_k(t)\|_2$,
\begin{equation}
\delta(t)=\frac{1}{N-1}\sum_{k=1}^{N-1}\frac{|\hat\ell_k(t)-\ell_0|}{\ell_0},
\end{equation}
reported as the mean over the rollout. Let $d_{jk}(t)$ be the shortest
distance between the centerlines of segments $j$ and $k$ and $D$ the cable
diameter; a pair with $|j-k|>2$, beyond the two-edge bending stencil, violates
when $d_{jk}(t)<D$, and the self-intersection rate is the mean over the rollout
of $v(t)=\mathbf{1}[\min_{|j-k|>2}d_{jk}(t)<D]$. For the phase split a
predicted frame is floor contact when $\min_i z_i(t)\le D$ on the recorded
trajectory after the input frame and free flight otherwise; each rollout
contributes the mean of $\rel(t)$ over its selected frames, and a phase cell is
the equal-weight mean over rollouts with at least one selected frame ($400$
floor contact and $345$ free flight of the $400$ windows). The per-seed window
standard deviations and counts are taken from the same per-window scores.

\end{document}